\documentclass[preprint,12pt]{elsarticle}

\usepackage{amssymb}
\usepackage{amsmath}
\usepackage{graphicx}
\usepackage{url}
\usepackage{hyperref}

\journal{Knowledge-Based Systems}

\begin{document}

\begin{frontmatter}

\title{Closing the Gap: Data-Centric Fine-Tuning of Vision Language Models for the Standardized Exam Questions}

\author[1]{Egemen Sert} 

\author[1,2]{Şeyda Ertekin}

\affiliation[1]{
  organization={Department of Computer Engineering, Middle East Technical University (METU)},
  city={Ankara},
  country={Türkiye}
}

\affiliation[2]{
  organization={METU-DTX Digital Transformation \& Innovation Centre, METU},
  city={Ankara},
  country={Türkiye}
}

\begin{abstract}
Multimodal reasoning has become a cornerstone of modern AI research. Standardized exam questions offer a uniquely rigorous testbed for such reasoning, providing structured visual contexts and verifiable answers. While recent progress has largely focused on algorithmic advances such as reinforcement learning (e.g., GRPO, DPO), the data-centric foundations of vision–language reasoning remain less explored.

We show that supervised fine-tuning (SFT) with high-quality data can rival proprietary approaches. To this end, we compile a 161.4-million-token multimodal dataset combining textbook question–solution pairs, curriculum-aligned diagrams, and contextual materials, and fine-tune Qwen-2.5VL-32B using an optimized reasoning syntax (QMSA). The resulting model achieves 78.6\% accuracy, only 1.0\% below Gemini 2.0 Flash, on our newly released benchmark YKSUniform, which standardizes 1,854 multimodal exam questions across 309 curriculum topics.

Our results reveal that data composition and representational syntax play a decisive role in multimodal reasoning. This work establishes a data-centric framework for advancing open-weight vision–language models, demonstrating that carefully curated and curriculum-grounded multimodal data can elevate supervised fine-tuning to near–state-of-the-art performance.
\end{abstract}

\begin{graphicalabstract}
\includegraphics[width=\textwidth]{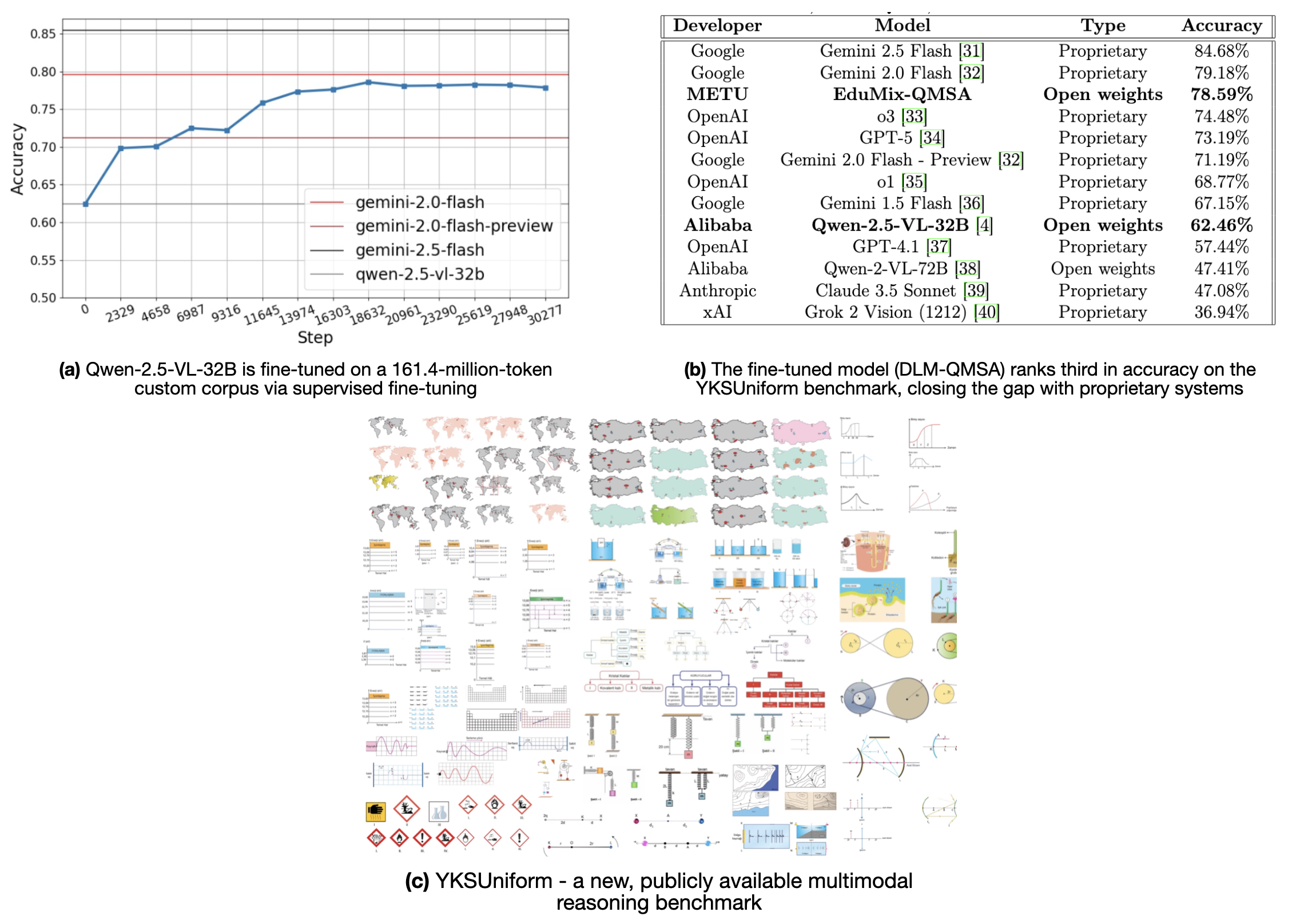}
\end{graphicalabstract}

\begin{highlights}
\item Demonstrates that high-quality, data-centric supervised fine-tuning (SFT) can substantially improve multimodal reasoning—approaching proprietary model performance without reinforcement learning.
\item Introduces a 161.4 million token multimodal corpus composed of three structured datasets: \textbf{CoreReason}, \textbf{MetaReason}, and \textbf{ContextVQA}, each addressing complementary reasoning dimensions.
\item Shows that structured dataset mixing outperforms single-source training, confirming the value of curriculum-aligned and multimodal supervision.
\item Identifies the \textbf{QMSA} syntax (\texttt{<question> <meta> <solution> <answer>}) as the optimal reasoning format, revealing that verbose teacher reasoning tokens can hinder generalization.
\item Releases \href{https://huggingface.co/datasets/egemensert/yksuniform}{\textbf{YKSUniform}}, a standardized benchmark of 1,854 multimodal Turkish high-school exam questions, and presents the open-weight \textbf{EduMix-QMSA} model achieving 78.6\% accuracy—only 1.0\% below Gemini 2.0 Flash.
\end{highlights}


\end{frontmatter}

\section{Introduction}
\label{sec:intro}

The ability of artificial intelligence systems to perform reasoning has gained tremendous momentum in recent years. Large language models (LLMs) and their successors are now explicitly optimized not only for generation but for \emph{structured reasoning}—chaining intermediate steps, validating sub-goals, and producing verifiable solutions. In parallel, a major shift has occurred toward \emph{multimodal reasoning}, where models integrate visual and textual inputs to solve more complex tasks that require perception and cognition jointly. Benchmark datasets such as MMMU~\cite{yue2024mmmu}, MathVista~\cite{lu2023mathvista}, and MathVision~\cite{wang2024measuring} exemplify this trend, challenging models to reason over images, diagrams, and text in a unified framework. Standardized exam questions, in particular, provide an ideal testbed for such reasoning: they are diverse in modality, demand multi-step logical inference, and yield objectively verifiable answers.

Vision–Language Models (VLMs) have evolved rapidly from perception-oriented systems into architectures capable of structured reasoning and cognitive processing. Early developments such as LLaVA~1.5~\cite{liu2024improvedbaselinesvisualinstruction} and KOSMOS-2~\cite{peng2023kosmos2groundingmultimodallarge} demonstrated how multimodal alignment between image encoders and text decoders could bridge the semantic gap between what is seen and what is written. Streamlined designs like Fuyu-8B~\cite{fuyu-8b} further removed dedicated encoders by directly projecting image patches into language tokens. As models matured, they began tackling reasoning-heavy benchmarks (e.g., MMMU~\cite{yue2024mmmu}, MathVista~\cite{lu2023mathvista}, MathVision~\cite{wang2024measuring}) and introducing explicit grounding to visual elements to improve factual consistency. Recent architectures such as Skywork~R1V2~\cite{wang2025skyworkr1v2multimodalhybrid}, Phi-4~\cite{abdin2024phi4technicalreport}, and Qwen-2.5-Omni~\cite{xu2025qwen25omnitechnicalreport} integrate multiple modalities—including text, vision, and even audio—via modular adapters or hybrid encoders, emphasizing both efficiency and interpretability. These developments mark a transition from mere visual description to genuine multimodal reasoning, enabling mathematical, scientific, and spatial problem-solving in open-world contexts.

VLMs have also expanded into traditional computer-vision tasks with new, generalized architectures. Models like PaliGemma~\cite{beyer2024paligemmaversatile3bvlm} and PaliGemma 2~\cite{steiner2024paligemma2familyversatile} can perform detection and segmentation via localization tokens, while Molmo~\cite{deitke2024molmopixmoopenweights} extends these capabilities to object counting and referencing. Qwen-2.5-VL~\cite{bai2025qwen25vltechnicalreport} further adapts these principles to interface understanding, broadening multimodal applications to agentic workflows. The performance of such models, however, remains heavily dependent on the quality and structure of their training data. While Supervised Fine-Tuning (SFT) on high-quality curated datasets forms the backbone of current VLM pipelines, the community has increasingly emphasized algorithmic improvements—most notably reinforcement learning from human feedback (RLHF)~\cite{ouyang2022traininglanguagemodelsfollow} and its efficient variants such as Proximal Policy Optimization (PPO)~\cite{schulman2017proximalpolicyoptimizationalgorithms} and Group Relative Policy Optimization (GRPO)~\cite{shao2024deepseekmathpushinglimitsmathematical}. These methods align models with human preferences and improve reasoning quality, but often require substantial computational resources. Simpler SFT-based approaches, exemplified by the s1~\cite{muennighoff2025s1simpletesttimescaling} and OpenThoughts~\cite{guha2025openthoughtsdatarecipesreasoning} datasets, have shown that data quality and structure alone can yield substantial reasoning improvements—yet their potential for multimodal reasoning remains underexplored.

Parallel innovations have targeted the reasoning process itself. Mixture-of-Experts architectures such as Kimi-VL~\cite{kimiteam2025kimivltechnicalreport} and Seed1.5-VL~\cite{guo2025seed15vltechnicalreport} scale reasoning capacity efficiently, while mechanisms like “Forced Rethinking” in VL-Rethinker~\cite{wang2025vlrethinkerincentivizingselfreflectionvisionlanguage} and the “Thinker–Talker” paradigm in Qwen-2.5-Omni~\cite{xu2025qwen25omnitechnicalreport} encourage explicit self-correction and reflective thought. Reinforcement-learning-based fine-tuning (e.g., GRPO~\cite{shao2024deepseekmathpushinglimitsmathematical}) now complements SFT to improve generalization, addressing challenges such as vanishing advantages and alignment tax~\cite{chu2025sftmemorizesrlgeneralizes}. Related advances—ranging from reasoning over knowledge graphs via reinforcement learning~\cite{chen2022rulemining,liu2022dynamic} to multimodal knowledge-graph embeddings~\cite{tran2025mesn} and path-based multi-hop reasoning~\cite{cui2023pathbased}—illustrate a clear trajectory: reasoning in AI is evolving toward systems that can plan, verify, and self-reflect across modalities.

Despite these algorithmic advances, the field remains strongly \emph{algorithm-oriented}, with limited understanding of the data principles that enable strong reasoning behavior. Open-weight models continue to trail proprietary systems, largely because multimodal reasoning datasets are fragmented, low-resource, or domain-specific. In particular, \emph{curriculum-grounded reasoning tasks}—those involving structured educational taxonomies, explicit metadata, and verifiable solutions—remain under-represented in current research. Standardized examinations naturally embody these characteristics, making them ideal for studying how model reasoning depends on data composition, representational syntax, and curriculum alignment.

In this work, we adopt a data-centric perspective and investigate how the structure and composition of multimodal data affect reasoning performance. We show that high-quality \emph{Supervised Fine-Tuning (SFT)} on curated, curriculum-aligned datasets can elevate open-weight VLMs to near-proprietary performance without reinforcement learning. To this end, we curate a large-scale multimodal corpus totaling 161.4 million tokens and fine-tune Qwen-2.5-VL~\cite{bai2025qwen25vltechnicalreport} using optimized reasoning syntax. We further introduce \textbf{YKSUniform}, a publicly available benchmark comprising 1,854 multimodal questions across 309 topics drawn from standardized educational materials. Although derived from a national curriculum, YKSUniform serves as a general framework for evaluating multimodal reasoning on structured, verifiable, and multilingual data.

\paragraph{Contributions}
\begin{itemize}
    \item We demonstrate that \textbf{data-centric Supervised Fine-Tuning (SFT)} with carefully curated multimodal data can substantially close the gap between open-weight and proprietary vision–language models. Our approach achieves 78.6\% accuracy with Qwen-2.5-VL, only 1.0\% below Gemini 2.0 Flash, without reinforcement learning or proprietary supervision.
    \item To enable this, we construct a \textbf{161.4-million-token multimodal dataset} integrating textbook-derived question–solution pairs, curriculum-aligned diagrams, and contextual learning materials, forming a scalable and balanced resource for multimodal reasoning.
    \item We introduce \textbf{YKSUniform}, a publicly available, standardized benchmark of 1,854 exam-style questions across 309 curriculum topics, designed to evaluate multimodal reasoning under structured, verifiable conditions.
\end{itemize}

Together, these contributions establish a data-centric framework for advancing open-weight VLMs, showing that carefully curated, curriculum-aligned multimodal data can be as decisive as model scale or reinforcement learning in driving reasoning performance.

\section{Methodology and Experiments}
\label{sec:data_collection}

We curated three complementary datasets—\textbf{CoreReason (CR)}, \textbf{MetaReason (MR)}, and \textbf{ContextVQA (CV)}—to systematically examine how multimodal data composition affects reasoning in supervised fine-tuning (SFT). Together, they span the full high-school curriculum with reliable solutions and rich multimodal coverage, totaling 161.4M tokens. Each dataset targets a distinct gap: \textbf{CoreReason} containts reasoning traces distilled from exam preparation questions, \textbf{MetaReason} contributes scale and structured curriculum metadata, and \textbf{ContextVQA} broadens contextual grounding through auxiliary text–image supervision. This design allows us to test the hypothesis that \emph{data composition and structure} are as critical as model scale for reasoning generalization.

\subsection{CoreReason Dataset}
We collected questions from 32 educational books published by two public sources, extracting 34,621 items and distilling 32,767 solutions using Gemini 2.0 Flash~\cite{comanici2025gemini25pushingfrontier}. To suppress teacher-model failure modes (e.g., speculation or meta-commentary), we filtered distilled solutions using a curated list of 44 rejection keywords and standardized all accepted samples to the \texttt{<question> <think> <solution> <answer>} (QTSA) format:
\begin{itemize}
    \item \texttt{<question>} — the OCR-rendered prompt and textual content extracted from the image,
    \item \texttt{<think>} — the intermediate reasoning trace generated by the teacher model,
    \item \texttt{<solution>} — the essential derivation steps leading to the result,
    \item \texttt{<answer>} — the final choice, restricted to \texttt{A–E}.
\end{itemize}

The filtering yielded 29,287 clean triplets. We then manually reviewed the ten lowest-performing topics, identifying 2,358 malformed samples (8.05\%) with incorrect solution steps despite correct answers. For these, we regenerated up to 20 candidates per item, recovering 2,096 valid solutions. Using an internal dashboard, we also performed targeted human edits on 995 items, discarding five additional samples. The resulting partially reviewed set, denoted \textbf{CoreReason-Reviewed}, contains 29,005 high-quality training samples and 1,854 held-out test items. The held-out split corresponds to the \textbf{YKSUniform} benchmark, which we  \href{https://huggingface.co/datasets/egemensert/yksuniform}{publicly release}. This standardized benchmark enables consistent evaluation across models and experiments.

\subsection{Training on CoreReason}
We fine-tuned Qwen-2.5VL-7B on CoreReason using 5×H200 GPUs for three epochs (batch size 5, learning rate $10^{-5}$). The best checkpoint reached 71.09\% accuracy. Training on CoreReason-Reviewed, despite a smaller sample size, further improved accuracy to 71.84\% (+0.75 points), confirming that quality-controlled supervision yields more consistent reasoning behavior than raw scale.

\subsection{MetaReason Dataset}
We next identified a large public question bank of 23,105 items labeled with subject, unit, and objective metadata and accompanied by video explanations. For each question, we:
\begin{enumerate}
    \item Sampled eight candidate solutions with Gemini 2.5 Flash and accepted any passing the CoreReason phrase filter,
    \item If rejected, injected the corresponding video transcript and regenerated three new solutions,
    \item Excluded items that failed both steps.
\end{enumerate}

This pipeline yielded 20,417 accepted samples in step (1) and 1,482 more in step (2), totaling 21,899 items. Each entry includes explicit metadata between \texttt{<meta>} tokens, encoding the question’s curriculum context.  

\subsection{Sequential Training: CoreReason $\rightarrow$ MetaReason}
Starting from the best CoreReason-trained checkpoint, fine-tuning on MetaReason improved accuracy to 75.84\%. Applying a masked-completion strategy—randomly masking 20\% of completion tokens during training—further increased accuracy to 76.81\%. These improvements likely result from a combination of factors, including differences in teacher model quality, question distribution, and the addition of structured metadata, which together contribute to stronger reasoning alignment. Later experiments (Table~\ref{table:7b-syntax-experiments}) further support that incorporating \texttt{<meta>} tokens positively impacts performance, suggesting that curriculum metadata provides useful grounding when integrated into reasoning supervision.

\subsection{ContextVQA Dataset}
To supplement reasoning with background knowledge (e.g., named entities, geographic or biological diagrams), we scraped three high-school curriculum blogs through the following pipeline:
\begin{enumerate}
    \item Extract the main article body,
    \item Capture one paragraph before and after each imagei
    \item Render KaTeX equations to \LaTeX,
    \item Extract question/solution/answer triplets if present,
    \item Snapshot embedded slide decks.
\end{enumerate}

This process produced 1,681 markdowns, 1,706 contextual images, and 1,215 slide decks (30,531 pages), plus 4,925 native Q/S/A triplets. We augmented these by (a) generating 20,187 synthetic Q/A pairs, (b) captioning 1,706 images with context-aware prompts, and (c) describing 30,531 slides, yielding a 60.8K synthetic multimodal corpus.  

Training solely on ContextVQA produced weaker results ($\approx$52.32\%), but its inclusion later proved valuable for enriching background understanding when combined with CoreReason and MetaReason.

\subsection{Putting the Pieces Together: Dataset Mix Experiments}
To isolate data-composition effects, we trained Qwen-2.5VL-7B on all dataset combinations for one epoch and evaluated each model on YKSUniform. Table~\ref{table:7b-dataset-experiments} reports the results.

\begin{table}[!htbp]
\centering
\caption{Dataset mix experiments on Qwen-2.5VL-7B. Each run is trained for one epoch and evaluated on YKSUniform. The best performance is obtained by combining all datasets.}
\label{table:7b-dataset-experiments}
\resizebox{0.5\textwidth}{!}{\begin{tabular}{||c c||} 
\hline
\textbf{Dataset Mix} & \textbf{Accuracy} \\
\hline \hline
\textbf{CR+MR+CV} & \textbf{55.99\%} \\
\hline
CR+CV & 55.61\% \\
MR+CV & 55.39\% \\
MR+CR & 55.02\% \\
MR & 53.99\% \\
CR & 53.40\% \\
MR (No Video) & 53.34\% \\ 
\hline
\end{tabular}}
\end{table}

Three trends emerge:  
(1) \textbf{MR} performs better than other individual datasets;  
(2) \textbf{CV} alone is weaker but provides complementary context when combined;  
(3) \textbf{CR+MR+CV} yields the highest accuracy, confirming that multimodal reasoning benefits from diverse, curriculum-grounded supervision.  
The best result (\textbf{55.99\%}) demonstrates that combining curated reasoning data, structured metadata, and contextual visual knowledge maximizes reasoning generalization.

\subsection{Syntax Mix Experiments: Q, M, T, S, A}
Having identified the optimal dataset mix, we next examined how the \emph{structure of reasoning supervision} affects model performance. Each syntax combination consists of the following components:
\begin{itemize}
    \item \textbf{Q: Question} — textual prompt extracted via OCR and visual parsing;
    \item \textbf{M: Meta} — curriculum metadata (subject, topic, difficulty);
    \item \textbf{T: Think} — intermediate reasoning trace from the teacher model;
    \item \textbf{S: Solution} — concise derivation of the correct answer;
    \item \textbf{A: Answer} — the final multiple-choice option.
\end{itemize}

We denote each configuration using these letters (e.g., \textbf{QMTSA} includes all components, while \textbf{SA} includes only solution and answer). All models were trained on the CR+MR+CV dataset for one epoch using Qwen-2.5VL-7B, with \textbf{S} and \textbf{A} present in every configuration.

\begin{table}[!htbp]
\centering
\caption{Syntax mix experiments on Qwen-2.5VL-7B. Each letter denotes a syntax component: (Q)uestion, (M)eta, (T)hink, (S)olution, (A)nswer. Each run is trained for one epoch and evaluated on YKSUniform.}
\label{table:7b-syntax-experiments}
\resizebox{0.5\textwidth}{!}{\begin{tabular}{||c c||} 
\hline
\textbf{Syntax Mix} & \textbf{Accuracy} \\
\hline \hline
\textbf{QMSA} & \textbf{59.28\%} \\
QMTSA & 57.07\% \\
MTSA & 56.20\% \\
QSA & 55.77\% \\
QTSA & 55.12\% \\
TSA & 53.99\% \\
SA & 53.67\% \\
\hline
\end{tabular}}
\end{table}

Three key observations arise:  
(1) \textbf{Meta beats Think.} Adding curriculum metadata (\textbf{M}) consistently improves reasoning over including teacher reasoning traces (\textbf{T}). \textbf{QMSA} (no \texttt{<think>} tokens) achieves 59.28\%, outperforming \textbf{QMTSA} (57.07\%) by +2.21 points.  
(2) \textbf{Think can be detrimental.} Teacher traces often inject verbose or unstable reasoning steps, lowering generalization.  
(3) \textbf{Question + Meta is optimal.} Combining the problem statement with metadata provides the most informative signal for reasoning alignment.  

In summary, metadata provides curricular grounding that helps the model select appropriate reasoning schemas, while teacher traces may overfit to specific solution styles. This finding—that reasoning syntax is as important as data content—guides our final fine-tuning setup.

\subsection{Final Fine-Tuning: EduMix-QMSA Model}
\begin{figure}[!htbp]
\centering
\includegraphics[width=\textwidth]{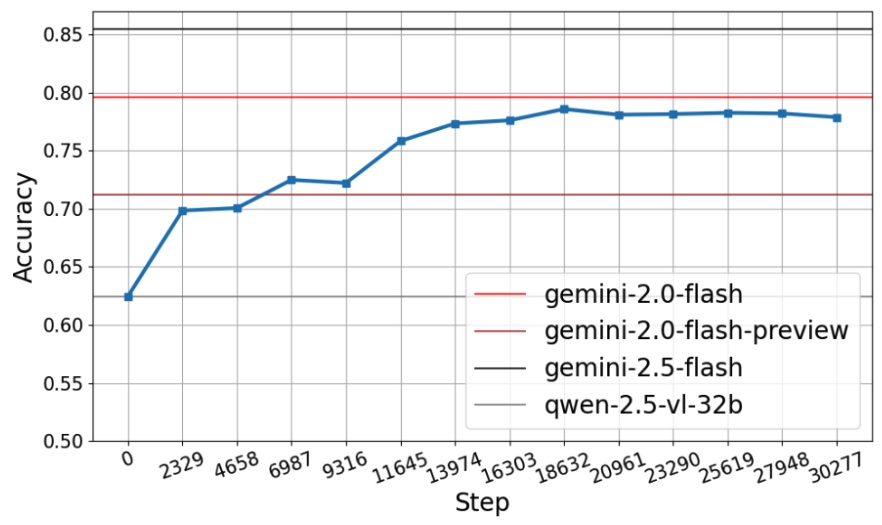}
\caption{Performance of our final model (EduMix-QMSA) trained using the CR+MR+CV QMSA scheme. The horizontal axis indicates training steps (one epoch = 4658 steps), and the vertical axis shows accuracy on YKSUniform. Performance of other models is shown for comparison.}
\label{fig:fin-performance}
\end{figure}

Having identified the optimal dataset composition (\textbf{CR+MR+CV}) and reasoning syntax (\textbf{QMSA}), we fine-tuned Qwen-2.5VL-32B using these configurations. We additionally employed masked language modeling~\cite{chen2024maskedthoughtsimplymasking}, randomly masking 20\% of completion tokens to improve generalization. Training ran for six epochs, with the best performance (78.59\%) achieved at epoch four (Figure~\ref{fig:fin-performance}). This represents a 25.8\% relative improvement over the baseline Qwen-2.5VL-32B model (62.46\%).

\begin{table}[!htbp]
\caption{Performance of open-weight and proprietary models on the YKSUniform dataset. Our fine-tuned model, EduMix-QMSA, ranks third overall.}
\label{table:benchmark}
\resizebox{\textwidth}{!}{\begin{tabular}{||c c c c||} 
\hline
\textbf{Developer} & \textbf{Model}                 & \textbf{Type}      & \textbf{Accuracy} \\
\hline \hline
Google    & Gemini 2.5 Flash~\cite{comanici2025gemini25pushingfrontier}  & Proprietary & 84.68\%      \\
Google    & Gemini 2.0 Flash~\cite{google2024geminiupdate}               & Proprietary & 79.18\%      \\
\textbf{METU} & \textbf{EduMix-QMSA} & \textbf{Open weights} & \textbf{78.59\%} \\
OpenAI    & o3~\cite{openai2025o3o4systemcard}        & Proprietary & 74.48\% \\
OpenAI    & GPT-5~\cite{openai2025gpt5systemcard}     & Proprietary & 73.19\% \\
Google    & Gemini 2.0 Flash - Preview~\cite{google2024geminiupdate} & Proprietary & 71.19\% \\
OpenAI    & o1~\cite{openai2024o1systemcard}          & Proprietary & 68.77\% \\
Google    & Gemini 1.5 Flash~\cite{geminiteam2024gemini15} & Proprietary & 67.15\% \\
\textbf{Alibaba} & \textbf{Qwen-2.5-VL-32B}~\cite{bai2025qwen25vltechnicalreport} & \textbf{Open weights} & \textbf{62.46\%} \\
OpenAI    & GPT-4.1~\cite{openai2024gpt4technicalreport} & Proprietary & 57.44\% \\
Alibaba   & Qwen-2-VL-72B~\cite{yang2024qwen2technicalreport} & Open weights & 47.41\% \\
Anthropic & Claude 3.5 Sonnet~\cite{anthropic2024claude35sonnet} & Proprietary & 47.08\% \\
xAI       & Grok 2 Vision (1212)~\cite{xai2024grok2} & Proprietary & 36.94\% \\
\hline
\end{tabular}}
\end{table}

This performance demonstrates that data-centric SFT with curriculum-aligned multimodal supervision can elevate open-weight VLMs to near-proprietary levels, underscoring the centrality of data design and representational syntax in multimodal reasoning.

\section{Conclusions}
Multimodal reasoning has emerged as a central challenge in artificial intelligence, where the interplay between visual understanding and structured textual reasoning defines model performance across domains. Recent advances have largely focused on algorithmic innovations, yet our findings highlight that \textbf{data-centric supervised fine-tuning} remains a powerful, underexplored driver of progress.

In this work, we showed that careful dataset composition and representational syntax can substantially narrow the performance gap between open-weight and proprietary vision–language models. By curating a 161.4M-token multimodal corpus and fine-tuning with the EduMix-QMSA configuration, we achieved near-proprietary performance on the YKSUniform benchmark—demonstrating that model capability can scale not only with size or reinforcement learning but also with \emph{data quality and structure}. 

Our contributions include the open release of \textbf{YKSUniform}, a standardized benchmark for multimodal educational reasoning, and the introduction of the \textbf{CoreReason}, \textbf{MetaReason}, and \textbf{ContextVQA} datasets for data-centric SFT research. Together, these resources establish a foundation for evaluating and improving reasoning models in structured, curriculum-grounded domains.

Recent work by Bansal et al. \cite{bansal2025honeybee} echoes our findings: caption-like grounding boosts multimodal reasoning, and overly long reasoning traces can hurt performance. While they report that heterogeneous data mixtures reduce accuracy, our aligned combination of CoreReason, MetaReason, and ContextVQA yields consistent gains, suggesting that synergy depends on shared reasoning style rather than diversity alone. Additionally, our use of explicit \texttt{<meta>} tokens - encoding subject, topic, and objective—further improves performance by conditioning the model on each question’s curricular context.

While our experiments focus on the Turkish high-school setting, the framework generalizes to any multilingual or low-resource educational context where multimodal reasoning is required. Future research may extend this approach to more abstract reasoning tasks, cross-lingual learning, and domain adaptation in data-scarce regions. Ultimately, this study underscores that progress in multimodal reasoning depends not only on algorithms but equally on the thoughtful design of the data that trains them.

\section{Acknowledgements}
This research received funding from the Research Universities Support Program (YOK-ADEP) with project number ADEP-312-2024-11490.

\bibliographystyle{elsarticle-num}

\end{document}